\documentclass[runningheads]{llncs}
\usepackage[T1]{fontenc}
\usepackage{graphicx}
\usepackage{amssymb}
\usepackage{amsmath}

\RequirePackage{xspace}
\makeatletter
\DeclareRobustCommand\onedot{\futurelet\@let@token\@onedot}
\def\@onedot{\ifx\@let@token.\else.\null\fi\xspace}

\def\ie{\emph{i.e}\onedot}

\def\etal{\emph{et al}\onedot}
\makeatother

\begin{document}
\title{Contrastive Gaussian Clustering: Weakly Supervised 3D Scene Segmentation}
\titlerunning{Contrastive Gaussian Clustering}
%
\author{Myrna C. Silva$^*$ \and
Mahtab Dahaghin$^*$ \and
Matteo Toso \and Alessio {Del Bue}}
\authorrunning{M.C. Silva and M. Dahaghin \etal}
%
\institute{Pattern Analysis and Computer Vision (PAVIS), Istituto Italiano di Tecnologia (IIT)\\
Genoa, Italy\\
\email{\{name.surname\}@iit.it}}
\maketitle              
\def\thefootnote{*}\footnotetext{These authors contributed equally to this work}\def\thefootnote{\arabic{footnote}}

\begin{abstract}
We introduce \textit{Contrastive Gaussian Clustering}, a novel approach capable of provide segmentation masks from any viewpoint and of enabling 3D segmentation of the scene. Recent works in novel-view synthesis have shown how to model the appearance of a scene via a cloud of 3D Gaussians, and how to generate accurate images from a given viewpoint by projecting on it the Gaussians before $\alpha$ blending their color. Following this example, we train a model to include also a segmentation feature vector for each Gaussian. These can then be used for 3D scene segmentation, by clustering Gaussians according to their feature vectors; and to generate 2D segmentation masks, by projecting the Gaussians on a plane and $\alpha$ blending over their segmentation features. Using a combination of contrastive learning and spatial regularization, our method can be trained on inconsistent 2D segmentation masks, and still learn to generate segmentation masks consistent across all views. Moreover, the resulting model is extremely accurate, improving the IoU accuracy of the predicted masks by $+8\%$ over the state of the art. 
Code and trained models will be released upon acceptance.
\end{abstract}

\section{Introduction}\label{sec:intro}


\begin{figure}[t]
    \includegraphics[width=\textwidth,trim={0cm 3cm 0cm 0cm}, clip]{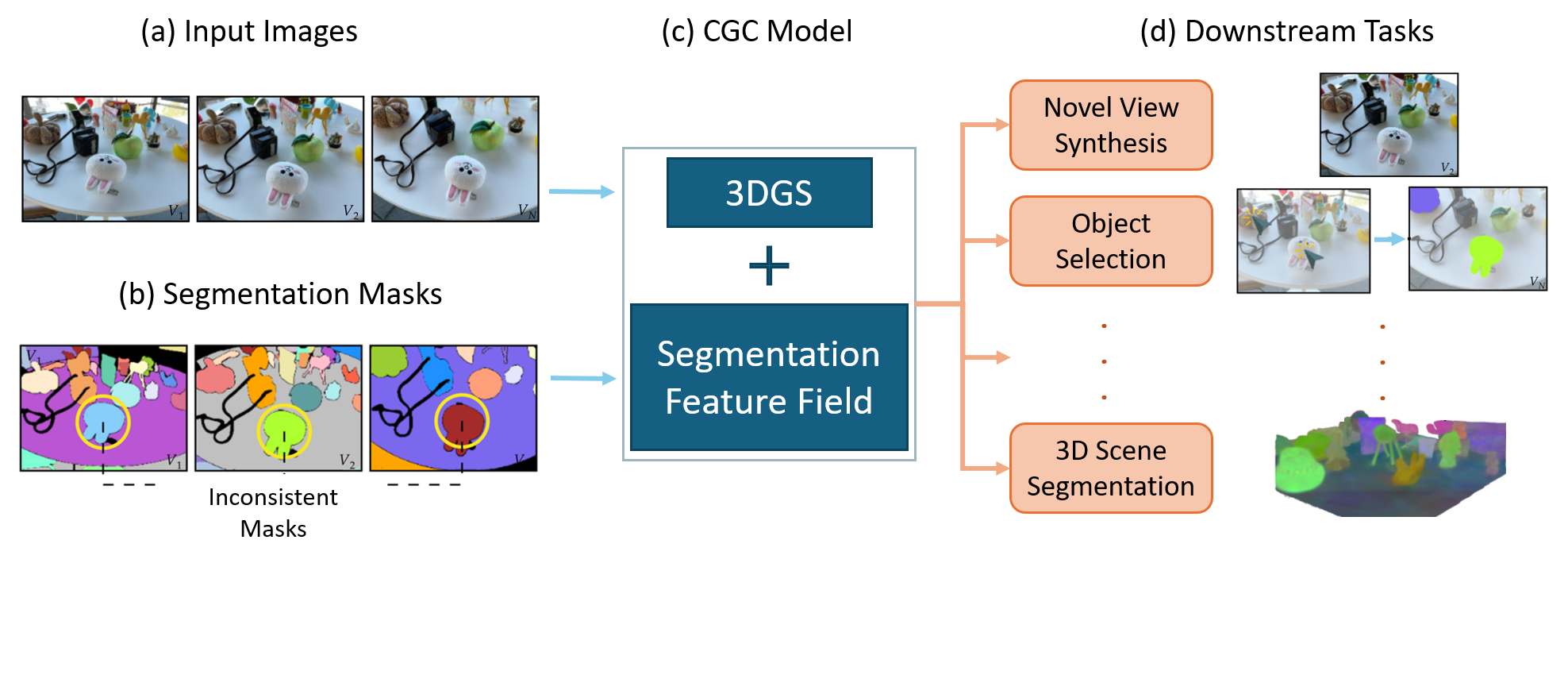}
    \caption{The objective of \textit{Contrastive Gaussian Clustering} is to take (a) a set of input images and (b) their independent segmentation masks and (c) distill their information in a model based on 3DGS. This model can then be used for (d) a wide range of visual and segmentation downstream tasks, such as novel view synthesis, retrieving the mask of a selected object, or 3D scene segmentation.
    }
    \label{fig:concept}
\end{figure}

Reliable and efficient 3D scene segmentation, \ie the ability of dividing the content of a 3D scene into separate objects, is a fundamental skill at the core of several computer vision tasks. It is a fundamental prerequisite for autonomous navigation, scene understanding and for many AR/VR applications~\cite{hou2019sis}.
In this work, we propose a general 3D scene segmentation approach based on 3D Gaussian Splatting~\cite{kerbl3Dgaussians}, applicable to any real-world scene and that only requires 2D images and their segmentation masks as input.

One of the challenges of 3D scene segmentation is the limited availability of annotated 3D scene datasets, as manual annotations are time-consuming~\cite{7785081}. In contrast, 2D image segmentation is nowadays readily available; as a consequence, recent works bypassed this issue by lifting 2D image understanding to 3D space~\cite{peng2023openscene}, inserting their semantic information into 3D point clouds~\cite{qi2017pointnet,qi2017pointnet++,hu2020randlanet} or NeRFs~\cite{lerf2023,Zhi:etal:ICCV2021,kunduPanopticFields}. These methods have shown that averaging noisy labels across multiple views allows for the generation of view-independent dense semantic labels~\cite{Zhi:etal:ICCV2021}. Early approaches relied on a limited range of task-specific labels~\cite{7298801,chen2020scanrefer}, but the recent introduction of foundational models 
like CLIP~\cite{radford2021learning} and SAM~\cite{kirillov2023segany} allows us to generate open-vocabulary 2D semantic segmentation labels, which can be used to optimize scene representations~\cite{qin2023langsplat,gaussian_grouping}. The segmentation masks generated by the foundation models, however, are not always consistent across views, and existing methods require 
time-consuming pre-processing to enforce cross-view consistency in the training data~\cite{gaussian_grouping}. In this work, we address this by introducing a model that can be trained on inconsistent 2D segmentation masks, while still learning a 3D feature field consistent across all views.

As exemplified in Figure~\ref{fig:concept}, our method takes as input a) a set of multi-view images and b) their 2D segmentation masks, which are not required to be consistent across views. We then use images and masks to train c) a model to represent both the visual and geometrical information of the scenes, as well as a 3D segmentation feature field. This model can then be used for a wide range of d) downstream tasks, based on the visual information (novel view synthesis), segmentation information (3D scene segmentation) or on a combination of them (returning a segmentation mask given a selected point on a rendered view).
The optimization of the geometric and visual component can be approached following standard 3D Gaussian Splatting~\cite{kerbl3Dgaussians}, using a rendering loss to optimize the color, position and shape of the 3D Gaussians. To learn the 3D segmentation feature field, we propose to extract information from the inconsistent 2D segmentation masks via contrastive learning. This approach ensures segmentation consistency across all views without requiring changes to the 2D masks themselves.
We test the proposed method against related works based on implicit scene representations~\cite{lerf2023} and 3D Gaussian representations~\cite{gaussian_grouping,qin2023langsplat}, and show through qualitative and quantitative evaluation how 
our method matches and outperforms them. A video outlining the motivation and main results of this paper is available in the Supplementary Material. 
Our contributions can be summarized as follows:
\begin{itemize}
\item An approach to embed a \emph{3D feature field} in a 3DGS model to allow novel view synthesis of segmentation features.
\item A contrastive-learning approach to enforce multi-view consistency in the semantic field, even when training on inconsistent segmentation masks.
\item An approach for 3D scene segmentation by clustering the Gaussians according to the feature field. 
\end{itemize}


\section{Related Work}
In this section, we provide an overview of the relevant literature on image and scene segmentation, besides 3D scene modeling with techniques for novel-view synthesis. For a complete review of scene understanding or semantic segmentation, we refer the readers to~\cite{8573760} and~\cite{garciagarcia2017review}, respectively.

\paragraph{\textbf{Scene Understanding}}
Scene understanding is a fundamental problem in computer vision, inferring the semantics and properties of all elements in a 3D scene given a 3D model and a set of RGB images~\cite{peng2023openscene}. Early approaches train the models on ground-truth (GT) 3D labels, focusing on specific tasks like 3D object classification~\cite{7298801}, object detection and localization~\cite{chen2020scanrefer} or 3D semantic and instance segmentation~\cite{Behley2019SemanticKITTIAD,Matterport3D,dai2017scannet,9786676}. To overcome the limited availability of 3D GT data, subsequent work leverage 2D supervision, by back-projecting and fusing 2D labels to generate pseudo 3D annotations~\cite{Genova2021Learning3S} or applying contrastive learning between 2D and 3D features~\cite{Liu2021ContrastiveMF,Sautier2022ImagetoLidarSD}. More recently, large visual language models~\cite{caron2021emerging,kirillov2023segany,radford2021learning} have allowed to shift from a close-set of predefined labels to an open-vocabulary framework~\cite{radford2021learning}, making possible zero-shot transfer to new tasks and dataset distributions. 
Like some of these works, we also leverage contrastive learning and foundation models, using class-agnostic segmentation masks generated by the Segment Anything Model (SAM)~\cite{kirillov2023segany}. However, we apply these techniques to a different scene representation - 3D Gaussian Splatting - and combine the contrastive loss with other forms of supervision, such as regularization based on the spatial distance between the Gaussians.


\paragraph{\textbf{Radiance Fields}}

Neural Radiance Fields (NeRF)~\cite{mildenhall2020nerf} optimize a Multilayer Perceptron (MLP) to represent a 3D scene as a continuous volumetric function that maps position and viewing direction to density and color. NeRF has enabled the rendering of complex scenes, producing high-quality results in novel-view synthesis. Subsequent work have focused on faster training/rendering~\cite{M_ller_2022,yu2021plenoxels,barronMIPNerf360}.
An alternative approach to NVS comes from 3D Gaussian Splatting (3DGS)~\cite{kerbl3Dgaussians}, which achieves both competitive training times and real-time rendering at higher image resolution. Unlike NeRF~\cite{mildenhall2020nerf} methodologies, 3DGS foregoes a continuous volumetric representation and instead approximate a scene using millions of 3D Gaussians with different sizes, orientations, and view-dependent colors. 
One of the advantages of this approach is that it allows for direct access to the radiance field data, enabling to edit the scene by removing, displacing or adding Gaussians 
~\cite{chen2024survey}. This also enable capturing dynamic scenes, including a time parameter to model the scene's changes over time~\cite{wu20234dgaussians}; or to combine the model in a pipeline with a foundation model to edit the scene from text prompts~\cite{GaussianEditor} or select the Gaussians associated with a specific object~\cite{gaussian_grouping}. Of these methods, Gaussian Grouping is the closest to our application by segmenting the scene into groups of 3D Gaussians. However, this technique relies on a video-tracker to obtain consistent masks IDs across the training images, which also preset the number of instances in the scene.


\paragraph{\textbf{Scene Understanding in Radiance Fields Representations}}

Semantic-NeRF~\cite{Zhi:etal:ICCV2021} extends the implicit scene representation to encoding appearance, geometry, and semantics. It has been suggested that Semantic-NeRF generates denoised semantic labels by training over sparse or noisy annotations. Another class of methods propose to distill image embeddings extracted by a foundation model encoder into a 3D feature field. Distilled Feature Fields (DFF)~\cite{kobayashi2022distilledfeaturefields} includes an extra branch that outputs a pixel-aligned feature vector extracted from LSeg~\cite{li2022languagedriven} or DINO~\cite{caron2021emerging}. Unlike DFF, LERF~\cite{lerf2023} supervises by rendering non pixel-aligned multi-scale CLIP~\cite{radford2021learning} embeddings. Although these techniques locate a wide variety of objects given any language prompt, they may suffer from inaccurate segmentations occasionally caused by objects with similar semantics.
More recent methods~\cite{qin2023langsplat,cen2023segment} have used foundational models for grounding language/segmentation features onto the 3D Gaussians. While these methods provide better performance in localization tasks, achieving higher accuracy, 
their segmentation masks are noisy/patchy. Mingqiao \textit{et al.}~\cite{gaussian_grouping} cluster the Gaussians by assigning them a unique identity ID. Though these methods can include instance segmentation features into the scene representation, the number of objects in the scene is predefined, and it requires an additional tracking method to pre-compute needed multi-view consistent segmentation labels. A similar approach of using contrastive learning to lift inconsistent 2D segmentations into NeRF has also been used for 3D instance segmentation~\cite{bhalgat2023contrastive}. We show an alternative to encode identity features into 3D Gaussians, so we can group them into clusters that we can easily extract/remove from the 3D scene.

\section{Methodology} \label{sec:method}

In this work, we represent a scene as a collection of 3D Gaussians that jointly encode geometry, appearance, and instance segmentation information. Our approach can render high-quality images and provide segmentation masks in real-time.
We draw inspiration from the 3DGS, which represents a static scene as a collection of estimated 3D Gaussians. We empower a 3DGS model to tackle 3D scene understanding downstream tasks by creating a 3DGS instance representation. To achieve this, we augment each 3D Gaussian with a feature vector that encodes instance information, used to segment a scene into distinct clusters. A comparison between our algorithm and 3DGS is available in the the Supp.Mat..





\begin{figure}[tb!]
    \centering
    \includegraphics[width=\textwidth, trim={0.1cm 0.1cm 0.1cm 0.1cm}, clip] {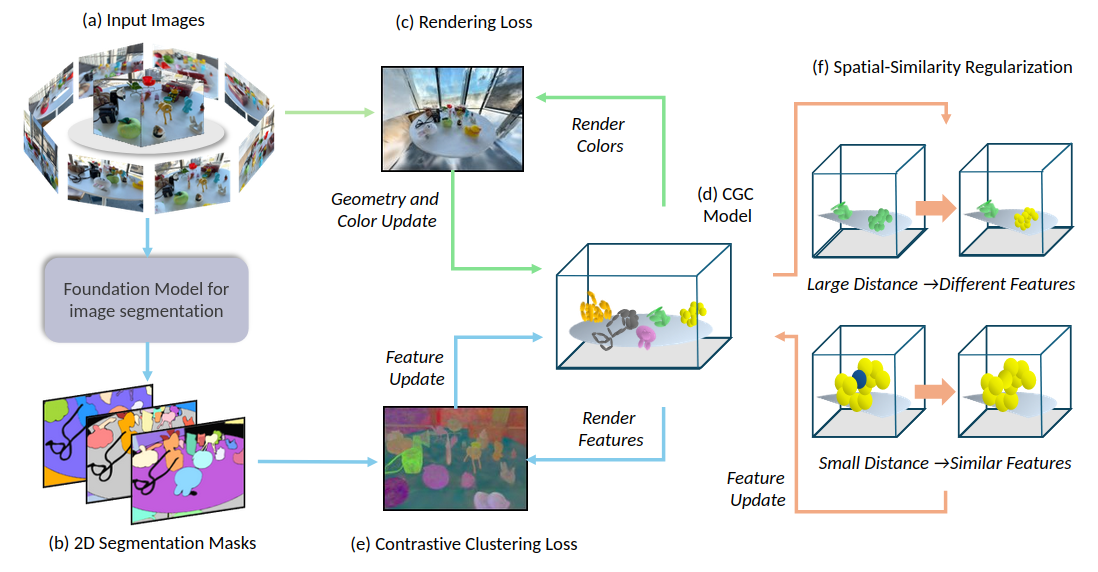}
    \caption{\textbf{Pipeline}: (a) Given a set of images from different viewpoints, we use (b) a foundation model for image segmentation to generate 2D segmentation masks. We capture the appearance of the scene via (c) a rendering loss that, like in traditional 3DGS, optimizes the geometry and color of (d) our \textit{Contrastive Gaussian Clustering} model. 
    Simultaneously, (e) a contrastive loss on the rendered-features learns a 3D segmentation feature field, which is also encoded in (d) our scene model. Moreover, we use (f) a spatial-similarity regularization mechanism, encouraging the segmentation features to be similar for neighboring Gaussians and different for faraway Gaussians.
    }
    \label{fig:pipeline}
\end{figure}

As shown in Figure~\ref{fig:pipeline}, our approach takes (a) a set of input images, from which we independently extract (b) 2D inconsistent segmentation masks using a foundation model for image segmentation. Then, we optimize the 3D Gaussians using (c) the original 3DGS loss function~\cite{kerbl3Dgaussians} that measures the difference between the rendered and ground truth images. Simultaneously, we make use of a (e) contrastive segmentation loss to supervise the 3D feature field. This results in a (d) 3D Gaussian scene representation which captures both visual and instance information. Our 3D feature field encapsulates the relation between 2D masks and the 3D scene. To provide more accurate segmentations and speed up training, we introduce (f) a regularization term that enforces a correlation between the distance of Gaussians in Euclidean and the segmentation feature space. 

In this section, we first review the 3DGS rendering method. Then we discuss the main steps of our pipeline, including rendering a 3D feature field and supervising the features via contrastive learning.

\subsection{Preliminaries on 3D Gaussian Splatting} \label{section:3DGS}
The 3DGS model represents a scene as millions of 3D Gaussians parameterized by their position $\mu$, 3D covariance matrix $\Sigma$, opacity $\alpha$, and color $c$. 3DGS represents the view-dependent appearance $c$ by spherical harmonics. These parameters are jointly optimized to render high-quality novel-views. Since 3DGS preserves the properties of differentiable volumetric representations, it requires only a set of images along with their corresponding camera parameters as input. Initially, 3DGS creates a set of 3D Gaussians using a sparse SfM point cloud obtained during camera calibration. To render 3D Gaussians from a particular point of view, the 3DGS starts by projecting these Gaussians onto the image space, a rendering termed ``splatting''. Subsequently, 3DGS generates a sorted list of $\mathcal{N}$ Gaussians, ordering them from closest to farthest. The color of a pixel $C$ is then computed by $\alpha$-blending the colors of $\mathcal{N}$ overlapping points:
\begin{equation}
    C = \sum_{i \in \mathcal{N}} c_i\alpha'_i\prod_{j=1}^{i-1}(1 - \alpha'_{j}). \label{eq:color}
\end{equation} 
The final opacity $\alpha'_i$ is determined by multiplying the learned opacity $\alpha_i$ and the 2D Gaussian. The optimization is done by subsequent iterations that compare the ground truth images against the corresponding rendered views.

\subsection{3D Feature Field} \label{section:3DFF}

The 3D feature field is a collection of learnable-vectors stored on the 3D Gaussians, that encode the instance segmentation of the scene. We augment each 3D Gaussian with a learnable feature $f$. Unlike the view-dependent appearance, 
this feature must remain consistent across all viewing directions. Therefore, instead of computing spherical harmonics coefficients, we directly extract its component from Gaussians. During training, we start with randomly initialized feature vectors. The 3D feature field optimization process involves two iterative steps repeated for each training view: rendering; and clustering rendered-features. 



At each iteration, we render an image and its corresponding 2D feature map, following an analogous process to the rendering algorithm described in Section~\ref{section:3DGS}. For each pixel of the desired view, we $\alpha$ blend the features as: 
\begin{equation}
    F = \sum_{i \in \mathcal{N}} f_i\alpha'_i\prod_{j=1}^{i-1}(1 - \alpha'_{j})
\end{equation}

\subsubsection{Contrastive Clustering} \label{section:contrastive-clustering-2d}


As the first step toward scene optimization, SAM automatically generates 2D segmentation masks from the set of input images. 
Specifically, we deploy SAM's automatic generation pipeline on each training image $I \in \mathbb{R}^{H \times W}$, resulting in sets of segments $\{m^p \in \mathbb{R}^{H \times W} | p = 1 \dots \mathcal{N}_k\}$. The number of segments per image $\mathcal{N}_k$ is uncapped, allowing the proposed approach to learn as many instances as are present in the scene.

As described previously, the 3D feature field optimization is composed of two steps that are iteratively repeated: rendering from a particular point of view a 2D feature map, and clustering its rendered-features based on the corresponding 2D segmentation mask to compute a contrastive clustering loss.

The core idea of contrastive clustering is maximizing the similarity among features within the same segment while minimizing for those from different segments. The cluster $\{f^p\}$, is the set of rendered-features that belongs to the same segment $m^p$. We compute the centroid $\Bar{f}^p$ as the mean feature in $\{f^p\}$. Our objective is to find features that minimize the following loss function:

\begin{equation}
    \mathcal{L}_{CC} = - \frac{1}{\mathcal{N}_k}\sum_{p=1}^{\mathcal{N}_k}\sum_{q=1}^{|\{f^p\}|} \log{\frac{\exp{\left( f_q^p \cdot \Bar{f}^p \mathbin{/} \phi^p \right)}}{\sum_{s=1}^{\mathcal{N}_k} \exp{\left( f_q^p \cdot \Bar{f}^s \mathbin{/} \phi^s \right)}}},
\end{equation}
where, $f_q^p$ are features in $\{f^p\}$. The temperature of the $p$-th cluster is $\phi^p$. Similar to \cite{ying2023omniseg3d}, we set the cluster temperature as:
$\sum_{q=1}^{\mathcal{N}_p} \lVert f_q^p - \Bar{f}^p \lVert_2 \mathbin{/} \mathcal{N}_p\log\left(\mathcal{N}_p+\epsilon \right)$, where $\mathcal{N}_p = |\{f^p\}|$ and $\epsilon=100$. Rather than regularize the features by including a normalization loss, we apply $\ell_2$-normalization to each feature in the rendered feature map before the loss computation.

\subsubsection{Spatial-Similarity Regularization} \label{section:contrastive-clustering-multiview}

An easy way to obtain the instance segmentation is to cluster similar features. However, we occasionally observe sparse outliers (Gaussians misclassified) in regions where the scene is not well observed. Furthermore, we notice that constant failures in the 2D segmentation (for instance, a chair that frequently is segmented in two parts: legs and seat) may induce to inaccurate segmentation masks. 

To address these issues, we include spatial-similarity regularization to enforce spatial continuity of the feature vectors, encouraging adjacent 3D Gaussians to have similar segmentation feature vectors while discouraging faraway Gaussians from having the same segmentation features. 
The regularization function is computed with $\mathcal{M}$ sampling Gaussians:

\begin{equation}
    \mathcal{L}_{regularization} = \frac{\lambda_{near}}{\mathcal{M}\mathcal{K}} \sum_{j=1}^{\mathcal{M}} \sum_{i=1}^{\mathcal{K}} H \left( 1 - f_j \cdot f_i \right ) + \frac{\lambda_{far}}{\mathcal{M}\mathcal{L}} \sum_{j=1}^{\mathcal{M}} \sum_{i=1}^{\mathcal{L}} H \left ( f_j \cdot f_i \right ),
\end{equation}
where $H$ denotes the sigmoid function. We compute the cosine similarity of features in the closest $\mathcal{K}=2$ and the farthest $\mathcal{L}=5$ Gaussians. We set $\lambda_{near} = 0.05$ and $\lambda_{far} = 0.15$. 

\subsubsection{Loss Function} \label{section:total-loss} 
Given the losses defined in this section, we train the model with the total loss:
\begin{equation}\label{totalloss}
    \mathcal{L} = \mathcal{L}_{rendering} + \lambda_{clustering}\mathcal{L}_{CC} + \mathcal{L}_{regularization},
\end{equation}
where $\mathcal{L}_{rendering}$ is the original rendering loss of 3DGS.
Empirically, we set $\lambda_{clustering}=1 \times 10^{-6}$.

\section{Experiments}

We aim to segment objects within a scene into distinct clusters, to mainly generate novel segmentation masks from any viewpoint of the scene. We therefore compare our algorithm against recent work for scene understanding, which code has already been published. 
Specifically, we compare our approach against three relevant competitors: LERF~\cite{lerf2023}, Gaussian Grouping~\cite{gaussian_grouping}, and LangSplat~\cite{qin2023langsplat}.
LERF is an open-vocabulary localization method that embeds a language field within a NeRF by grounding CLIP embeddings extracted at multiple scales over the training images. Given a text query, LERF predicts 3D regions with the semantic content pertinent to the input query. The recent Gaussian Grouping~\cite{gaussian_grouping} is a technique for classifying 3D Gaussians into predefined instances and LangSplat~\cite{qin2023langsplat} is an approach that results in a collection of 3D Language Gaussians, such as LERF, outputs a relevancy map for a given text.
We evaluate performance using two metrics: the mean intersection over union (mIoU), which measures the overlap of the GT and rendered masks; and the mean boundary intersection over union (mBIoU), which evaluates contour alignment between predicted and ground truth masks. In both cases, we report the average performance over all test views and text prompts. 

In this section, we first provide details about the datasets used to evaluate the models (Section~\ref{sec.exp.dataset}), then we give some implementation details (Section~\ref{sec.exp.impelemnt}) and report the segmentation performance of the models (Section~\ref{sec.exp.results}). Finally, we discuss the advantages of a spatial-similarity regularization loss (Section~\ref{sec.exp.ablation}).


\subsection{Datasets}\label{sec.exp.dataset}
We evaluate the chosen models on two datasets, containing indoor and outdoor scenes: the LERF-Mask dataset~\cite{gaussian_grouping} and the 3D-OVS dataset~\cite{liu2023weakly}.
\paragraph{\textbf{LERF-Mask}} The LERF-Mask dataset is composed of three manually annotated scenes from the LERF-Localization dataset~\cite{lerf2023} dataset. These scenes belong to the ``posed long-tailed objects'' of LERF-Localization, which are scenes containing multiple objects with low search volume and low competition, arranged on a plane, like a set of objects arranged on a small table (``Figurines'').
These scenes are captured using the Polycam application on an iPhone, utilizing its onboard SLAM to obtain the camera poses.
\paragraph{\textbf{3D-OVS}} 
We also report quantitative and qualitative results on five scenes of the 3D-OVS dataset~\cite{liu2023weakly}, which also consists of a set of long-tail objects, such as toys and everyday objects on a ``Bed'' or on a ``Sofa''.


\subsection{Implementation Details} \label{sec.exp.impelemnt}
The models evaluated in this section are supervised on segmentation masks automatically generated with the ViT-H SAM model, trained on the SA-1B dataset~\cite{kirillov2023segany}. These masks are used to learn feature vectors in $\mathbb{R}^{16}$ for each Gaussian. To ensure a stable training process, the loss terms of Eq.(~\ref{totalloss}) are applied with different frequencies: the standard 3DGS loss, used to optimize the geometrical and appearance aspects of the scene, is used at every training iteration. The contrastive clustering loss every $50$ iterations, and the spatial-similarity regularization every $100$ iterations. Moreover, to reduce the size of the problem and make the loss more stable, we evaluate the clustering loss only on clusters composed by more than $100$ features.
The optimization of a single scene takes approximately 30k iterations on an NVIDIA 4090 GPU, which amounts to approximately 20 minutes. The trained model then can render a novel segmentation mask in $0.005$ seconds; comparing this with the time necessary to run ViT-H SAM on an image ($5.1$ sec), this highlights the advantage of the proposed method.

\subsubsection{Instance segmentation} \label{subsubsec:instance_seg}
After optimization, the model can be used for \textit{Object Selection}, as exemplified in Fig.~\ref{fig:concept}; given one calibrated image, we want to find the segmentation mask associated with a given selected pixel.
Given a 2D pixel location in the image, we obtain a discriminative feature, \ie the rendered feature vector at that pixel's location. We then generate a 2D similarity map $S_C$ by rendering segmentation features for all pixels of the image, and evaluating their cosine similarity to the discriminative vector. 
Each pixel of the view $(u, v) \in I$ is then categorized as part of the object of interest or not. Pixels with cosine similarity greater than a fixed threshold $t$ (empirically, chosen as $t = 0.7$) are classified as part of the object; otherwise, they are not. The segmentation mask $M_{OBJ}$ is defined as: 

\begin{equation} \label{eq:segmentation_mask}
    M_{OBJ}(u, v) = \begin{cases} 
    \mbox{1,} & \mbox{if } S_C(u, v) \geq t \\ 
    \mbox{0,} & \mbox{otherwise}
    \end{cases}
\end{equation}

We note that this process can be applied in parallel to multiple objects, by extracting a set of discriminative features at different locations. An analogous approach also allows the 3D segmentation of the scene, by selecting one or more Gaussians and extracting, for each, all Gaussians with a high similarity score.

\subsubsection{Semantic Segmentation}
The proposed model renders novel feature maps by projecting and blending the content of the 3D feature field on an image plane. To compare these against the ground truth mask, we follow this procedure:
\emph{i)} we select a text prompt related to the content of the scene; \emph{ii)} we feed into Grounding DINO~\cite{liu2023grounding} an \texttt{(Image, Text)} pair which provides a bounding box that we use to generate a segmentation mask by using it as a prompt to SAM; \emph{iii)} we sample the rendered feature map associated to a pixel within the segment, and \emph{iv)} use it as a discriminative feature, generating the object's segmentation in an arbitrary view by selecting all pixels whose rendered-feature vector is falls within a predefined threshold from the discriminative feature. 

\subsection{Evaluation on Features}\label{sec.exp.results}

First, we compare the performance of our Contrastive Gaussian Clustering against its competitors. We report the average performance on each scene, but a complete breakdown of the performance on each object is available in the Supp.Mat.. 

\begin{table}[th!]
    \centering
    \caption{Comparison of semantic segmentation on LERF-Mask dataset. We report the mIoU and mBIoU (higher is better). LERF-Mask dataset contains accurate segmentation masks that we use to evaluate our segmentation performance.}
    \begin{tabular}{c|cc|cc|cc|cc}
        {} & \multicolumn{2}{c|}{Figurines} & \multicolumn{2}{c|}{Ramen} & \multicolumn{2}{c|}{Teatime} & \multicolumn{2}{c}{Average} \\
        Method & mIoU & mBIoU & mIoU & mBIoU & mIoU & mBIoU & mIoU & mBIoU \\
        \hline
        LERF~\cite{lerf2023} & 33.5 & 30.6 & 28.3 & 14.7 & 49.7 & 42.6 & 37.2 & 29.3 \\
        Gaussian Grouping~\cite{gaussian_grouping} & \underline{69.7} & \underline{67.9} & \textbf{77.0} & \textbf{68.7} & \underline{71.7} & \underline{66.1} & \underline{72.8} & \underline{67.6} \\
        LangSplat~\cite{qin2023langsplat} & 44.3 & 41.9 & 34.8 & 28.7 & 54.3 & 48.8 & 44.5 & 39.8 \\
        Ours & \textbf{91.6} & \textbf{88.8} & \underline{68.7} & \underline{63.1} & \textbf{80.5} & \textbf{78.9} & \textbf{80.3} & \textbf{76.9} \\
    \end{tabular}
    
    \label{tab:quantitative_results}
\end{table}

\begin{figure}
    \centering
    \includegraphics[width=\textwidth]{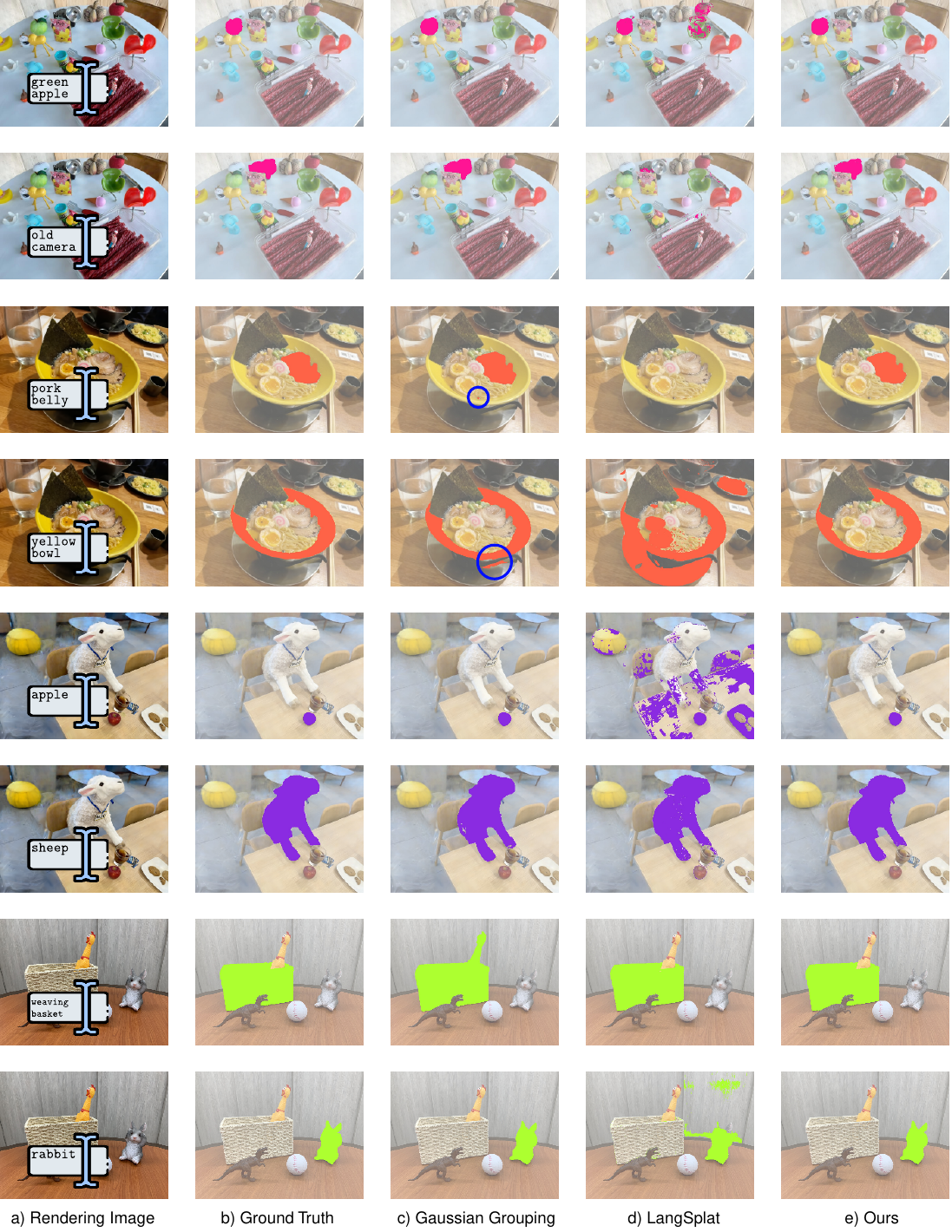}
    \caption{Qualitative comparison of test views for scenes on LERF-Mask dataset. Our method is able to generate accurate instance segmentation masks for any object on in-the-wild scenes. We replicate and exceed the results in \textit{green apple}, \textit{pork belly}, \textit{apple}. LangSplat exhibits noisy segmentation mask for \textit{old-camera} and coarse segmentation for \textit{sheep}. Gaussian Grouping misclassified some pixels outside \textit{yellow bowl} and \textit{pork belly} (marked with a blue circle) or classify two objects in the same category in \textit{waving basket}.}
    \label{fig:qualitative}
\end{figure}

As shown in Table~\ref{tab:quantitative_results}, our method significantly outperforms the other approaches on both metrics, providing on average a $+43 \%$ accuracy than LERF, $+36 \%$ accuracy than LangSplat, and $+8 \%$ accuracy than Gaussian Grouping on average. Regarding the masks' boundary quality, we outperform on average the competitors by $48 \%$, $37 \%$, and $9 \%$. Though Gaussian Grouping achieves better performance on \textit{Ramen}, we suggest seeing Fig.~\ref{fig:qualitative}, in which we show how our method produces better qualitative results, with more accurate segmentations.

\begin{table}[th!]
    \centering
    \caption{Comparison of semantic segmentation on 3D-OVS dataset, on scenes with sparse long-tail objects and simple background. We report the mIoU (higher is better).} 
    
    \resizebox{\textwidth}{!}{ \setlength{\tabcolsep}{0.55em} \begin{tabular}{c|c|c|c|c|c|c}
        {} & {Bed} & {Bench} & {Room} & {Sofa} & {Lawn} & {Average} \\
        Method & mIoU & mIoU& mIoU & mIoU & mIoU & mIoU \\
        \hline
        LERF~\cite{lerf2023} & 73.5 & 53.2 & 46.6 & 27.0 & 73.7 & 54.8 \\
        Gaussian Grouping~\cite{gaussian_grouping} & \textbf{97.3} & 73.7 & \underline{79.0} & \textbf{68.1} & \textbf{96.5} & \underline{82.9} \\
        LangSplat~\cite{qin2023langsplat} & 34.3 & \underline{84.8} & 56.3 & \underline{67.7} & \underline{95.8} & 67.8 \\
        Ours & \underline{95.2} & \textbf{96.1} & \textbf{86.8} & 67.5 & 91.8 & \textbf{87.5} \\
    \end{tabular}}
    
    \label{tab:quantitative_results_3D-OVS}
\end{table}

When we test the models on the 3D-OVS, the performance are comparable with the previous experiments, as shown in Table~\ref{tab:quantitative_results_3D-OVS}. In this case, we achieve the best performances in two out of five scenes, outperforming the competitors by $+20 \%$ and $+5 \%$ on average. 
In three out of the five scenes, Gaussian Grouping attains better numerical results; however, as shown in Figure~\ref{fig:qualitative}, in a qualitative comparison our methods achieves better instance segmentation.

Of the competitor models, Gaussian Grouping is the one that achieves the closes performance to us. The main limitation of this method is that, while it also enforces multi-view consistency, it does so through preprocessing, requiring that the 2D segmentation masks are made consistent. However, errors in this process propagate to the model, resulting in worse performance. In contrast, our model is not affected by this problem, as it autonomously learns to enforce consistency across the various views. The limited performance of LangSplat are instead due to its embedding in the image semantic features, embedded as 3-dimensional vector, without having a mechanism to ensure no two segments have similar features; this results in noisy segmentation masks and misdetections. This does not happen in our method, since the contrastive clustering loss ensures features from different segments are far in feature space.

Finally, Figure~\ref{fig:qualitative} provides a qualitative comparison of the methods. We can see that the resulting segmentation masks are compatible with the numerical results, showing how our method produces qualitative better instance segmentations than our competitors. Additional results showing the qualitative performance on 3D segmentation 
is available in the Supp.Mat.. 


\subsection{Ablation Studies}\label{sec.exp.ablation}

\begin{figure}[tb!]
    \centering
    \includegraphics[width=\textwidth]{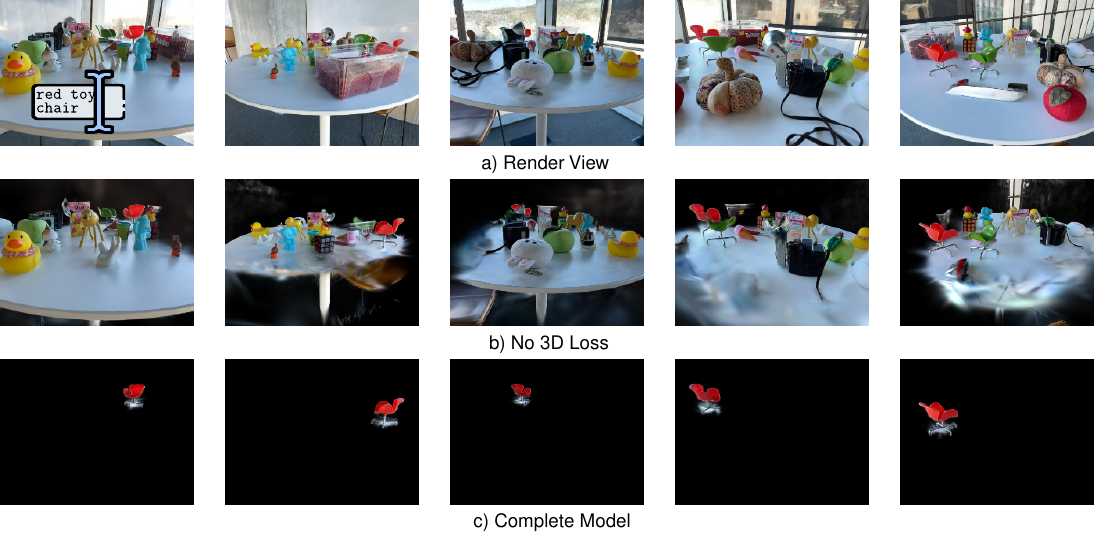}
    \caption{In this experiment, we extract the Gaussians that belong to the \textit{red toy chair}. We first compute a discriminative feature, following the same procedure as described in Section~\ref{subsubsec:instance_seg}. Then, we filter the 3D Gaussians by computing its similarity score. 
    The final result is the 3D segmentation of the \textit{red toy chair}. Observe that without our spatial-similarity regularization loss, the 3D segmentation is affected by a high number of outliers. Though these outliers can be easily removed by modifying the similarity threshold, we point-out that the outliers for a fixing similarity threshold is minimum when we use our spatial-similarity regularization loss.}
    \label{fig:ablation}
\end{figure}

\begin{table}[th!]
    \centering
    \caption{An ablation study of our model. In this experiment, we study the impact of the spatial-similarity regularization loss on the segmentation quality. 
    Metrics are averaged over all the test views. }
    \resizebox{\textwidth}{!}{\begin{tabular}{c|cc|cc|cc|cc}
        {} & \multicolumn{2}{c|}{Figurines} & \multicolumn{2}{c|}{Ramen} & \multicolumn{2}{c|}{Teatime} & \multicolumn{2}{c}{Average} \\
        Method & mIoU & mBIoU & mIoU & mBIoU & mIoU & mBIoU & mIoU & mBIoU \\
        \hline
        No 3D Loss & \underline{91.7} & 88.4 & \underline{67.6} & \underline{62.4} & 77.2 & 73.9 & 78.8 & 74.9 \\
        No Similarity Loss & \textbf{92.2} & \textbf{89.2} & 67.1 & 62.2 & 75.3 & 73.9 & 78.2 & 75.1 \\
        No Dissimilarity Loss & 91.4 & 88.6 & 67.1 & 61.9 & \underline{80.0} & \underline{78.6} & \underline{79.5} & \underline{76.4} \\
        Complete Model & 91.6 & \underline{88.8} & \textbf{68.7} & \textbf{63.1} & \textbf{80.6} & \textbf{78.9} & \textbf{80.3} & \textbf{76.9} \\
    \end{tabular}}
    \label{tab:ablation}
\end{table}

In the previous experiments we have claimed that the advantage of our method and, to a lesser extent, of Gaussian Grouping on the other methods is due to the implicitly learned multi-view consistency; which, in our case, is enforced through the loss of Eq.(~\ref{totalloss}). To validate this assumption, we run an ablation test comparing the performance of our model with and without 
spatial-similarity regularization. The results, reported in Table~\ref{tab:ablation}, show that on most scenes the spatial-similarity loss results in a significant performance improvement, of on average $78.8 \%$ against $80.3 \%$. This is also supported by the qualitative results on 3D segmentation reported in Figure~\ref{fig:ablation}.


\section{Conclusions}

In this paper, we introduce Contrastive Gaussian Clustering, a novel approach for 3D scene segmentation. We have shown how, by explicitly enforcing a multi-view consistency loss, we are able to learn consistent segmentation features from an inconsistent set of 2D segmentation masks. This means that the proposed model can learn from automatically generated segmentation masks, with little to none preprocessing required. 
Moreover, the use of a contrastive loss ensures that the features learned for Gaussians corresponding to different 3D clusters are distinct enough to provide accurate 3D segmentation. 
The combination of such two losses results in an efficient and accurate model, that outperforms current approaches based both on NERF and 3DGS.

\paragraph{Limitations.}

While the results reported in the paper are very promising, including additional information involves some trade-off. Foremost, the use of the two additional losses involves a computational overhead with respect to standard 3DGS, requiring on average  $100 \%$ longer time to train. We can, however, reduce this by only applying the losses every 50/100 iterations, respectively. Moreover, the additional information stored in the Gaussians requires additional memory capacity, in the future we would like to consider another ways to include the identity information into the scene representation.
Another limitations are inherited from SAM and Grounding DINO. For example, to select all Gaussians matched to a given semantic label, we rely on Grounding DINO to select that object' location in a reference image.
However, if this location is wrong, it will not be possible to recover the correct mask. 
The model's performance is also limited by the accuracy of the 2D segmentations used in training. We observe that, if multiple views contain incorrect masks, this can result into multiple instances being clustered together.

\paragraph{Future Works.} 
We will expand the proposed approach, integrating it with LLM for language interaction, and extending the feature field to also include hierarchical segmentations. Future work will explore more advanced ways of contrastive clustering. Concerning our multi-view contrastive loss, in future work we could explore more intelligent ways to contrast all the feature-objects.

\bibliographystyle{splncs04}
\bibliography{main}

\begin{thebibliography}{10}
\providecommand{\url}[1]{\texttt{#1}}
\providecommand{\urlprefix}{URL }
\providecommand{\doi}[1]{https://doi.org/#1}

\bibitem{barronMIPNerf360}
Barron, J.T., et~al.: {Mip-NeRF 360}: Unbounded anti-aliased neural radiance fields. In: CVPR (2022)

\bibitem{Behley2019SemanticKITTIAD}
Behley, J., Garbade, M., Milioto, A., Quenzel, J., Behnke, S., Stachniss, C., Gall, J.: {SemanticKITTI}: A dataset for semantic scene understanding of lidar sequences. ICCV  (2019)

\bibitem{bhalgat2023contrastive}
Bhalgat, Y., Laina, I., Henriques, J.F., Zisserman, A., Vedaldi, A.: {Contrastive Lift}: 3{D} object instance segmentation by slow-fast contrastive fusion. In: NeurIPS (2023)

\bibitem{caron2021emerging}
Caron, M., Touvron, H., Misra, I., J{\'e}gou, H., Mairal, J., Bojanowski, P., Joulin, A.: Emerging properties in self-supervised vision transformers. In: ICCV (2021)

\bibitem{cen2023segment}
Cen, J., Fang, J., Yang, C., Xie, L., Zhang, X., Shen, W., Tian, Q.: {Segment Any 3D Gaussians}. arXiv preprint arXiv:2312.00860  (2023)

\bibitem{Matterport3D}
Chang, A., Dai, A., Funkhouser, T., Halber, M., Niessner, M., Savva, M., Song, S., Zeng, A., Zhang, Y.: {Matterport3D}: Learning from {RGB-D} data in indoor environments. 3DV  (2017)

\bibitem{chen2020scanrefer}
Chen, D.Z., Chang, A.X., Nie{\ss}ner, M.: {Scanrefer: 3D object localization in rgb-d scans using natural language}. In: ECCV (2020)

\bibitem{chen2024survey}
Chen, G., Wang, W.: {A Survey on 3D Gaussian Splatting} (2024)

\bibitem{dai2017scannet}
Dai, A., Chang, A.X., Savva, M., Halber, M., Funkhouser, T., Nie{\ss}ner, M.: {ScanNet}: Richly-annotated 3{D} reconstructions of indoor scenes. In: CVPR (2017)

\bibitem{GaussianEditor}
Fang, J., Wang, J., Zhang, X., Xie, L., Tian, Q.: {GaussianEditor}: Editing 3{D} gaussians delicately with text instructions. arXiv preprint arXiv:2311.16037  (2023)

\bibitem{garciagarcia2017review}
Garcia-Garcia, A., Orts-Escolano, S., Oprea, S., Villena-Martinez, V., Garcia-Rodriguez, J.: A review on deep learning techniques applied to semantic segmentation (2017)

\bibitem{Genova2021Learning3S}
Genova, K., Yin, X., Kundu, A., Pantofaru, C., Cole, F., Sud, A., Brewington, B., Shucker, B., Funkhouser, T.A.: Learning 3d semantic segmentation with only 2d image supervision. 3DV  (2021)

\bibitem{hou2019sis}
Hou, J., Dai, A., Nie{\ss}ner, M.: {3D-SIS: 3D Semantic Instance Segmentation of RGB-D Scans}. In: CVPR (2019)

\bibitem{hu2020randlanet}
Hu, Q., Yang, B., Xie, L., Rosa, S., Guo, Y., Wang, Z., Trigoni, N., Markham, A.: {Randla-Net}: Efficient semantic segmentation of large-scale point clouds. In: CVPR (2020)

\bibitem{7785081}
Hua, B.S., Pham, Q.H., Nguyen, D.T., Tran, M.K., Yu, L.F., Yeung, S.K.: {SceneNN: A Scene Meshes Dataset with aNNotations}. In: 3DV (2016)

\bibitem{kerbl3Dgaussians}
Kerbl, B., Kopanas, G., Leimk{\"u}hler, T., Drettakis, G.: {3D Gaussian Splatting for Real-Time Radiance Field Rendering}. ACM Transactions on Graphics  (2023)

\bibitem{lerf2023}
Kerr, J., Kim, C.M., Goldberg, K., Kanazawa, A., Tancik, M.: {LERF: Language Embedded Radiance Fields}. In: ICCV (2023)

\bibitem{kirillov2023segany}
Kirillov, A., Mintun, E., Ravi, N., Mao, H., Rolland, C., Gustafson, L., Xiao, T., Whitehead, S., Berg, A.C., Lo, W.Y., Doll{\'a}r, P., Girshick, R.: {Segment Anything}. arXiv:2304.02643  (2023)

\bibitem{kunduPanopticFields}
Kundu, A., et~al.: {Panoptic Neural Fields: A Semantic Object-Aware Neural Scene Representation}. In: CVPR (2022)

\bibitem{li2022languagedriven}
Li, B., Weinberger, K.Q., Belongie, S., Koltun, V., Ranftl, R.: {Language-driven Semantic Segmentation}. In: ICLR (2022)

\bibitem{9786676}
Liao, Y., Xie, J., Geiger, A.: {KITTI-360: A Novel Dataset and Benchmarks for Urban Scene Understanding in 2D and 3D}. TPAMI  (2023)

\bibitem{liu2023weakly}
Liu, K., Zhan, F., Zhang, J., Xu, M., Yu, Y., El~Saddik, A., Lu, S.: {Weakly Supervised 3D Open-vocabulary Segmentation}. In: NeurIPS (2023)

\bibitem{liu2023grounding}
Liu, S., Zeng, Z., Ren, T., Li, F., Zhang, H., Yang, J., Li, C., Yang, J., Su, H., Zhu, J., et~al.: {Grounding dino: Marrying dino with grounded pre-training for open-set object detection}. arXiv preprint arXiv:2303.05499  (2023)

\bibitem{Liu2021ContrastiveMF}
Liu, Y., Fan, Q., Zhang, S., Dong, H., Funkhouser, T.A., Yi, L.: Contrastive multimodal fusion with tupleinfonce. ICCV  (2021)

\bibitem{mildenhall2020nerf}
Mildenhall, B., Srinivasan, P.P., Tancik, M., Barron, J.T., Ramamoorthi, R., Ng, R.: {NeRF: Representing Scenes as Neural Radiance Fields for View Synthesis}. In: ECCV (2020)

\bibitem{M_ller_2022}
Müller, T., Evans, A., Schied, C., Keller, A.: Instant neural graphics primitives with a multiresolution hash encoding. ACM Transactions on Graphics  (2022)

\bibitem{8573760}
Naseer, M., Khan, S., Porikli, F.: Indoor scene understanding in 2.5/3d for autonomous agents: A survey. IEEE Access  (2019)

\bibitem{peng2023openscene}
Peng, S., Genova, K., Jiang, C.M., Tagliasacchi, A., Pollefeys, M., Funkhouser, T.: {OpenScene: 3D Scene Understanding with Open Vocabularies} (2023)

\bibitem{qi2017pointnet++}
Qi, C.R., et~al.: {PointNet++: Deep Hierarchical Feature Learning on Point Sets in a Metric Space}. In: NeurIPS (2017)

\bibitem{qi2017pointnet}
Qi, C., Su, H., Mo, K., Guibas, L.: Pointnet: Deep learning on point sets for 3d classification and segmentation. In: Proceedings of the IEEE Conference on Computer Vision and Pattern Recognition. pp. 652--660 (2017)

\bibitem{qin2023langsplat}
Qin, M., Li, W., Zhou, J., Wang, H., Pfister, H.: {LangSplat: 3D Language Gaussian Splatting} (2023)

\bibitem{radford2021learning}
Radford, A., Kim, J.W., Hallacy, C., Ramesh, A., Goh, G., Agarwal, S., Sastry, G., Askell, A., Mishkin, P., Clark, J., et~al.: Learning transferable visual models from natural language supervision. In: Proceedings of the 38th International Conference on Machine Learning. pp. 8748--8763. PMLR (2021)

\bibitem{Sautier2022ImagetoLidarSD}
Sautier, C., Puy, G., Gidaris, S., Boulch, A., Bursuc, A., Marlet, R.: {Image-to-Lidar Self-Supervised Distillation for Autonomous Driving Data}. CVPR  (2022)

\bibitem{kobayashi2022distilledfeaturefields}
{Sosuke Kobayashi and Eiichi Matsumoto and Vincent Sitzmann}: Decomposing nerf for editing via feature field distillation. In: NeuIPS (2022)

\bibitem{wu20234dgaussians}
Wu, G., Yi, T., Fang, J., Xie, L., Zhang, X., Wei, W., Liu, W., Tian, Q., Xinggang, W.: {4D Gaussian Splatting for Real-Time Dynamic Scene Rendering}. arXiv preprint arXiv:2310.08528  (2023)

\bibitem{7298801}
Wu, Z., Song, S., Khosla, A., Yu, F., Zhang, L., Tang, X., Xiao, J.: {3D ShapeNets: A deep representation for volumetric shapes}. In: CVPR (2015)

\bibitem{gaussian_grouping}
Ye, M., Danelljan, M., Yu, F., Ke, L.: {Gaussian Grouping: Segment and Edit Anything in 3D Scenes}. arXiv preprint arXiv:2312.00732  (2023)

\bibitem{ying2023omniseg3d}
Ying, H., Yin, Y., Zhang, J., Wang, F., Yu, T., Huang, R., Fang, L.: Omniseg3d: Omniversal 3d segmentation via hierarchical contrastive learning (2023)

\bibitem{yu2021plenoxels}
Yu, A., Fridovich-Keil, S., Tancik, M., Chen, Q., Recht, B., Kanazawa, A.: {Plenoxels: Radiance Fields without Neural Networks} (2021)

\bibitem{Zhi:etal:ICCV2021}
Zhi, S., Laidlow, T., Leutenegger, S., Davison, A.J.: In-place scene labelling and understanding with implicit scene representation. In: ICCV (2021)

\end{thebibliography}


\begin{thebibliography}{10}
\providecommand{\url}[1]{\texttt{#1}}
\providecommand{\urlprefix}{URL }
\providecommand{\doi}[1]{https://doi.org/#1}

\bibitem{qin2023langsplat}
Qin, M., Li, W., Zhou, J., Wang, H., Pfister, H.: {LangSplat: 3D Language Gaussian Splatting} (2023)

\bibitem{gaussian_grouping}
Ye, M., Danelljan, M., Yu, F., Ke, L.: {Gaussian Grouping: Segment and Edit Anything in 3D Scenes}. arXiv preprint arXiv:2312.00732  (2023)

\end{thebibliography}

\end{document}


\title{Supplementary Material – Contrastive Gaussian Clustering: Weakly Supervised 3D Scene Segmentation}
\titlerunning{Contrastive Gaussian Clustering}
%
\author{Myrna C. Silva$^*$ \and
Mahtab Dahaghin$^*$ \and
Matteo Toso \and Alessio {Del Bue}}
\authorrunning{M.C. Silva and M. Dahaghin \etal}
%
\institute{Pattern Analysis and Computer Vision (PAVIS), Istituto Italiano di Tecnologia (IIT)\\
Genoa, Italy\\
\email{\{name.surname\}@iit.it}}
\maketitle              
\def\thefootnote{*}\footnotetext{These authors contributed equally to this work}\def\thefootnote{\arabic{footnote}}



To convey a better understanding of the proposed approach, we now provide additional details on the results presented in the main paper. First, we provide implementation details on how the qualitative results for the competitor baselines were produced (Sec.~\ref{sec:baselineimplementation}). Then, we outline the differences between vanilla 3DGS and our approach by highlighting the differences in the respective algorithm (Sec.~\ref{sec:pseudocode}). To provide a deeper understanding of the quantitative results in the paper, we also provide a breakdown of the performance on all objects present in the LERF-Mask and 3D-OVS scenes (Sec.~\ref{sec:perobjectresults}). Finally, we present qualitative results to better highlight the accuracy of our approach on 3D segmentation (Sec.~\ref{sec:qualitativemvm}) and 2D segmentation mask prediction (Sec~\ref{sec:qualitativelast}).

\section{Additional Baseline Details}\label{sec:baselineimplementation}

In this section, we provide additional insights into the implementation details of the baseline methods Gaussian Grouping~\cite{gaussian_grouping} and LangSplat~\cite{qin2023langsplat}. For the quantitative evaluation of LERF-Mask, we reported the official results of Gaussian Grouping provided in the original paper. For its qualitative evaluation and to obtain a per-object breakdown, instead, we had to run the baseline method again in order to have a consistent evaluation, \ie to generate outputs for all baseline models and our approach from the same input image/ text prompt. 

\paragraph{\textbf{Gaussian Grouping~\cite{gaussian_grouping}}} To evaluate the Gaussian Grouping baseline, we use the open-source code released by the authors\footnote{\url{https://github.com/lkeab/gaussian-grouping}} and follow their instructions for training and evaluation. The Gaussian Grouping implementation requires a significant amount of GPU memory, which for some scenes exceeds the available resources. To address this, we resize the LERF-Mask training images for the \textit{Figurines} to an image resolution of $740 \times 546$. For the \textit{Teatime} scene, we resize the training data at $889 \times 657$. For the \textit{Ramen} scene, we use the original data with an image resolution of $986 \times 728 $ pixels. For the 3D-OVS dataset, we train Gaussian Grouping's models at full resolution ($1440 \times 1080$).

\paragraph{\textbf{LangSplat~\cite{qin2023langsplat}}} We use the LangSplat open-source code released by the authors\footnote{\url{https://github.com/minghanqin/LangSplat}} and follow their instructions to preprocess, train, and render on a single scene. We follow their implementation details described in~\cite{qin2023langsplat} to implement our evaluation code. Specifically, for each text query we derive three relevancy maps, from which we obtain three binary masks. Then, we compute the average relevancy under each mask and select the binary mask related to the highest relevancy score as the predicted segmentation result.

\section{Additional Optimization Details}\label{sec:pseudocode}

	\begin{algorithm}[h!]
		\caption{Contrastive Gaussian Clustering}
		\label{alg:optimization}
		\begin{algorithmic}
			\State $P \gets$ SfM Points	\Comment{Positions}
			\State $S, C, A, \color{red}F\color{black} \gets$ InitAttributes() \Comment{Covariances, Colors, Opacities, \textcolor{red}{Features}}
			\State $i \gets 0$	\Comment{Iteration Count}
			
			\While{not converged}
			
			\State $V, \hat{I}, \color{red}M\color{black} \gets$ SampleTrainingView()	\Comment{Camera $V$, Image and \textcolor{red}{Segmentation}}
			\State $I, \color{red}F_{2D}\color{black} \gets$ Rasterize($P$, $S$, $C$, $A$, $V$) \Comment{Image and \textcolor{red}{Feature Map}}

                \color{red}
                \State $L_{rendering} \gets Loss(I, \hat{I}) $ \Comment{Rendering Loss}
                
                \State $L_{CC} \gets Loss_{CC}(M, F_{2D}) $ \Comment{Contrastive Clustering Loss}

                \State $L_{regularization} \gets Loss_{SSR}(P, F) $ \Comment{Spatial-Similarity Regularization}

                \State $L \gets L_{rendering} + L_{CC} + L_{regularization}$
			\color{black}
                
                \State $P$, $S$, $C$, $A$, \color{red}$F$\color{black} $\gets$ Adam($\nabla L$) \Comment{Backprop \& Step}

			\If{IsRefinementIteration($i$)}
			\ForAll{Gaussians $(\mu, \Sigma, c, \alpha, \color{red}f\color{black})$ $\textbf{in}$ $(P, S, C, A, \color{red}F\color{black})$}
			\If{$\alpha < \epsilon$ or IsTooLarge($\mu, \Sigma)$}	\Comment{Pruning}
			\State RemoveGaussian()	
			\EndIf
			\If{$\nabla_p L > \tau_p$} \Comment{Densification}
			\If{$\|S\| > \tau_S$}	\Comment{Over-reconstruction}
			\State SplitGaussian($\mu, \Sigma, c, \alpha, \color{red}f\color{black}$)
			\Else								\Comment{Under-reconstruction}
			\State CloneGaussian($\mu, \Sigma, c, \alpha, \color{red}f\color{black}$)
			\EndIf	
			\EndIf
			\EndFor		
			\EndIf
			\State $i \gets i+1$
			\EndWhile
		\end{algorithmic}
	\end{algorithm}

The proposed Contrastive Gaussian Clustering approach is based on the standard 3DGS implementation. However, to introduce segmentation feature vectors and additional contrastive loss and spatial regulariziation loss, we implemented several changes in the \textit{optimization and densification algorithm} at the core of 3DGS. In Algorithm~\ref{alg:optimization} we provide an overview of the difference between the two algorithms, marking in red the components introduced in our work. 





\begin{table}[th!]
    \centering
    \caption{Per-object breakdown of the performance reported in Table $1$ of the main paper on the \textit{Figurines}, \textit{Ramen} and \textit{Teatime} scenes of LERF-Mask. We report the the mean intersection over union (mIoU) and boundary intersection over union (mBIoU).}
    \label{tab:lerfresults}
    \resizebox{\textwidth}{!}{\begin{tabular}{c|c|cc|cc|cc}
        \multicolumn{2}{c}{}& \multicolumn{2}{|c|}{Gaussian Grouping~\cite{gaussian_grouping}} & \multicolumn{2}{c|}{LangSplat~\cite{qin2023langsplat}} & \multicolumn{2}{c}{Ours} \\
        \multicolumn{2}{c|}{Query} & \hspace*{1.5mm} mIoU \hspace*{1.5mm} & \hspace*{1.5mm} mBIoU \hspace*{1.5mm} & \hspace*{1.5mm} mIoU \hspace*{1.5mm} & \hspace*{1.5mm} mBIoU \hspace*{1.5mm} & \hspace*{1.5mm} mIoU \hspace*{1.5mm} & \hspace*{1.5mm} mBIoU \hspace*{1.5mm} \\
        \hline
        \multirow{8}{*}{\STAB{\rotatebox[origin=c]{90}{Figurines}}}\hspace*{1.75mm} & green apple & \underline{90.6}\hspace*{1.5mm} & \underline{88.0} & 38.5 & 34.5 & \textbf{95.4} & \textbf{93.3} \\
        & porcelain hand & \underline{84.2} & \underline{84.1} & 32.5 & 32.5 & \textbf{88.2} & \textbf{87.9} \\
        & green toy chair &  \underline{85.8} & \underline{83.6} & 67.4 & 65.4 & \textbf{92.3} & \textbf{91.4} \\
        & rubber duck with red hat & \underline{67.9} & \underline{67.6} & 31.4 & 32.1 & \textbf{93.0} & \textbf{92.9} \\
        & red toy chair & \underline{88.3} & \underline{86.3} & 75.6 & 74.4 & \textbf{93.0} & \textbf{91.8} \\
        & old camera & \underline{83.6} & \underline{78.5} & 22.3 & 19.5 & \textbf{85.0} & \textbf{78.9} \\
        & red apple & \underline{62.2} & \underline{54.5} & 42.8 & 35.1 & \textbf{94.1} & \textbf{85.2} \\
        \cline{2-8}
        & Average & \underline{80.4} & \underline{77.5} & 44.3 & 41.9 & \textbf{91.6} & \textbf{88.8} \\
    \hline
        \multirow{7}{*}{\STAB{\rotatebox[origin=c]{90}{Ramen}}} & chopsticks & \textbf{83.9} & \textbf{83.9} & 40.6 & 40.6 & \underline{83.7} & \underline{83.7} \\
        & pork belly & \underline{94.4} & \underline{83.2} & 0.0 & 0.0 & \textbf{96.5} & \textbf{88.3} \\
        & wavy noodles in bowl & \textbf{16.7} & \underline{11.8} & \underline{15.2} & \textbf{12.6} & 0.1 & 0.1 \\
        & yellow bowl & \underline{91.1} & \underline{83.0} & 61.3 & 55.7 & \textbf{94.3} & \textbf{88.6} \\
        & glass of water & 83.3 & \underline{66.0} & \underline{85.2} & 60.5 & \textbf{85.4} & \textbf{69.3} \\
        & egg &  \textbf{91.1} & \textbf{81.4} & 6.6 & 2.7 & \underline{52.0} & \underline{48.7} \\
        \cline{2-8}
        & Average & \textbf{76.8} & \textbf{68.2} & 34.8 & 28.7 & \underline{68.7} & \underline{63.1} \\
    \hline
        \multirow{11}{*}{\STAB{\rotatebox[origin=c]{90}{Teatime}}} & apple & \underline{90.4} & \underline{89.7} & 2.9 & 2.9 & \textbf{94.1} & \textbf{94.0} \\
        & plate & \underline{87.3} & \underline{87.3} & 72.7 & 72.7 & \textbf{89.6} & \textbf{89.6} \\
        & sheep & \underline{94.9} & \underline{80.1} & 77.4 & 50.8 & \textbf{97.1} & \textbf{90.4} \\
        & coffee mug & \underline{82.0} & \underline{82.0} & 67.4 & 67.4 & \textbf{84.9} & \textbf{84.9} \\
        & bag of cookies & \underline{92.7} & \underline{92.2} & 69.2 & 69.2 & \textbf{95.4} & \textbf{95.3} \\
        & paper napkin & \underline{84.7} & \underline{84.7} & 75.9 & 75.8 & \textbf{93.8} & \textbf{93.4} \\
        & spoon handle & 0.0 & 0.0 & \underline{0.4} & \underline{0.4} & \textbf{57.6} & \textbf{57.6} \\
        & cookies on a plate & \underline{22.3} & \underline{14.6} & \textbf{35.2} & \textbf{35.2} & 0.6 & 0.6 \\
        & tea in a glass & \underline{81.3} & \underline{77.5} & 77.6 & 74.0 & \textbf{94.8} & \textbf{93.0} \\
        & stuffed bear & 63.1 & \underline{43.7} & \underline{64.5} & 40.0 & \textbf{97.4} & \textbf{90.3} \\
        \cline{2-8} 
        & Average & \underline{69.9} & \underline{65.2} & 54.3 & 48.8 & \textbf{80.5} & \textbf{78.9} \\
    \hline
    \end{tabular}}
\end{table}

\begin{table}[th!]
    \centering
    \caption{Per-object breakdown of the performance reported in Table $1$ of the main paper on the \textit{Bed}, \textit{Bench}, \textit{Room}, \textit{Sofa} and \textit{Lawn} scenes of 3D-OVS. We report the the mean intersection over union (mIoU) and boundary intersection over union (mBIoU).}
    \label{tab:ovsresults}
    \resizebox{\textwidth}{!}{\begin{tabular}{c|c|cc|cc|cc}
         \multicolumn{2}{c}{}& \multicolumn{2}{|c|}{Gaussian Grouping~\cite{gaussian_grouping}} & \multicolumn{2}{c|}{LangSplat~\cite{qin2023langsplat}} & \multicolumn{2}{c}{Ours} \\
        \multicolumn{2}{c|}{Query} &  \hspace*{1.5mm} mIoU \hspace*{1.5mm} & \hspace*{1.5mm} mBIoU \hspace*{1.5mm} & \hspace*{1.5mm} mIoU \hspace*{1.5mm} & \hspace*{1.5mm} mBIoU \hspace*{1.5mm} & \hspace*{1.5mm} mIoU \hspace*{1.5mm} & \hspace*{1.5mm} mBIoU \hspace*{1.5mm} \\
        \hline
        \multirow{7}{*}{\STAB{\rotatebox[origin=c]{90}{Bed}}}\hspace*{1.75mm} & banana & \textbf{96.9} & \textbf{94.6} & 17.8 & 12.1 & \underline{95.9} & \underline{92.2} \\
        & black leather shoe & \underline{97.6} & \underline{92.9} & 71.4 & 49.3 & \textbf{97.9} & \textbf{94.1} \\
        & camera & \textbf{95.3} & \textbf{89.6} & 3.3 & 3.7 & \underline{85.0} & \underline{68.1} \\
        & hand & \textbf{96.8} & \textbf{91.3} & 6.9 & 3.2 & \underline{95.5} & \underline{88.5} \\
        & red bag & \textbf{98.5} & \textbf{93.0} & 81.2 & 43.6 & \underline{98.1} & \underline{78.5} \\
        & white sheet & \textbf{99.0} & \textbf{94.6} & 25.5 & 18.0 & \textbf{99.0} & \underline{89.5} \\
        \cline{2-8} 
        & Average & \textbf{97.3} & \textbf{92.7} & 34.3 & 21.7 & \underline{95.2} & \underline{85.2} \\
    \hline
        \multirow{8}{*}{\STAB{\rotatebox[origin=c]{90}{Bench}}} & pebbled concrete wall & \textbf{98.7} & \textbf{91.9} & 42.6 & 18.8 & \textbf{98.7} & \underline{84.6} \\
        & wood & \underline{97.8} & \underline{85.8} & 88.0 & 60.7 & \textbf{98.0} & \textbf{89.9} \\
        & Portuguese egg tart & \underline{95.4} & \underline{94.0} & 87.4 & 84.9 & \textbf{97.2} & \textbf{96.3} \\
        & orange cat & 94.5 & 87.5 & \underline{96.4} & \underline{92.6} & \textbf{97.3} & \textbf{93.4} \\
        & green grape & 0.0 & 0.0 & \underline{93.7} & \underline{87.4} & \textbf{95.5} & \textbf{88.5} \\
        & mini offroad car & \textbf{93.7} & 91.3 & \textbf{93.7} & \textbf{92.0} & 93.6 & \underline{91.8} \\
        & dressing doll & 36.0 & 36.5 & \underline{92.1} & \underline{86.9} & \textbf{92.2} & \textbf{87.3} \\
        \cline{2-8} 
        & Average & 73.7 & 69.6 & \underline{84.8} & \underline{74.8} & \textbf{96.1} & \textbf{90.3} \\
    \hline
        \multirow{7}{*}{\STAB{\rotatebox[origin=c]{90}{Room}}} & wood wall & \textbf{99.4} & \textbf{96.7} & 24.3 & 12.1 & \underline{43.6} & \underline{49.1} \\
        & shrilling chicken & 0.0 & 0.0 & \textbf{94.7} & \textbf{94.1} & \underline{91.4} & \underline{90.5} \\
        & weaving basket & 87.9 & 68.4 & \textbf{98.3} & \textbf{85.1} & \underline{97.4} & \underline{80.8} \\
        & rabbit & \underline{96.3} & \underline{92.2} & 45.3 & 37.5 & \textbf{97.0} & \textbf{94.6} \\
        & dinosaur & \underline{92.7} & \underline{92.1} & 6.8 & 6.1 & \textbf{94.9} & \textbf{94.5} \\
        & baseball & \textbf{97.5} & \textbf{95.5} & 68.4 & 63.1 & \underline{97.4} & \underline{94.6} \\
        \cline{2-8} 
        & Average & \underline{79.0} & \underline{74.2} & 56.3 & 49.7 & \textbf{86.9} & \textbf{84.0} \\
    \hline
        \multirow{7}{*}{\STAB{\rotatebox[origin=c]{90}{Sofa}}} & pikachu & \underline{48.8} & \underline{41.7} & \textbf{89.6} & \textbf{79.8} & 0.0 & 0.0 \\
        & a stack of UNO cards & \textbf{95.8} & \textbf{91.5} & 79.6 & 63.4 & \underline{95.4} & \underline{87.0} \\
        & a red Nintendo Switch & \multirow{2}{*}{\underline{90.9}} & \multirow{2}{*}{\underline{86.4}} & \multirow{2}{*}{89.8} & \multirow{2}{*}{84.9} & \multirow{2}{*}{\textbf{92.6}} & \multirow{2}{*}{\textbf{87.4}} \\
        & joy-con controller & & & & & & \\
        & Gundam & 16.1 & 21.2 & \textbf{79.5} & \textbf{73.3} & \underline{76.9} & \underline{70.8} \\
        & Xbox wireless controller & 59.9 & 52.0 & \underline{67.8} & \underline{57.0} & \textbf{96.1} & \textbf{86.7} \\
        & grey sofa & \textbf{97.4} & \textbf{81.8} & 0.0 & 0.0 & \underline{44.0} & \underline{46.0} \\
        \cline{2-8} 
        & Average & \textbf{68.1} & \underline{62.4} & \underline{67.7} & 59.7 & 67.5 & \textbf{63.0} \\
    \hline
        \multirow{6}{*}{\STAB{\rotatebox[origin=c]{90}{Lawn}}} & red apple & \textbf{96.3} & \textbf{95.1} & \underline{94.0} & 89.9 & 93.8 & \underline{91.0} \\
        & New York Yankees cap & \textbf{98.4} & \underline{95.1} & \textbf{98.4} & \textbf{95.4} & 97.9 & 88.5 \\
        & stapler & 94.7 & \underline{92.4} & \textbf{96.2} & \textbf{93.8} & \underline{95.6} & 90.3 \\
        & black headphone & \textbf{94.5} & \textbf{90.7} & \underline{91.7} & \underline{87.4} & 70.2 & 61.6 \\
        & hand soap & \textbf{96.0} & \textbf{93.5} & \underline{95.2} & \underline{91.3} & 93.7 & 90.7 \\
        & green lawn & 99.2 & \underline{94.0} & \textbf{99.5} & \textbf{96.2} & \underline{99.3} & 77.1 \\
        \cline{2-8} 
        & Average & \textbf{96.5} & \textbf{93.5} & \underline{95.8} & \underline{92.3} & 91.8 & 83.2 \\
    \hline
    \end{tabular}}
\end{table}

\section{Additional Details on the Quantitative Results}\label{sec:perobjectresults}
In the main paper, we showed how our method outperformed the available baselines on two datasets, LERF-Mask and 3D-OVS (Table $1$ and $2$ respectively). In this section, we provide additional insight by reporting the performance on each text query. 
The performance on the LERF-Mask dataset is reported in Table~\ref{tab:lerfresults}, and the ones on 3D-OVS in Table~\ref{tab:ovsresults}. 
First, we point out how the per-object results are consistent with the average results reported in the main paper,indicating the consistency of our model's performance across objects. Additionally, the breakdown of performance by objects provides additional insight into cases in which Gaussian Grouping outperforms our model, like in the \textit{Ramen} LERF-Mask scene. In this case, we outperform our principal competitor in three out of six objects. However, due to incorrect localization by Grounding DINO, we fail to correctly localize the segmentation of \textit{wavy noodles in bowl} and \textit{egg}, leading to significant errors. Similarly, in the \textit{Sofa} scene of the 3D-OVS dataset, we fail to find Pikachu. As a final remark, we point out that on 3D-OVS while we provide comparable performance to Gaussian Grouping in terms of Intersection over Union, overall we generate masks with lower boundary error, leading to better mBIoU performance.

\section{Additional Details in Downstream Tasks}\label{sec:qualitativemvm} 

\begin{figure}[thb!]
    \centering
    \includegraphics[width=\textwidth]{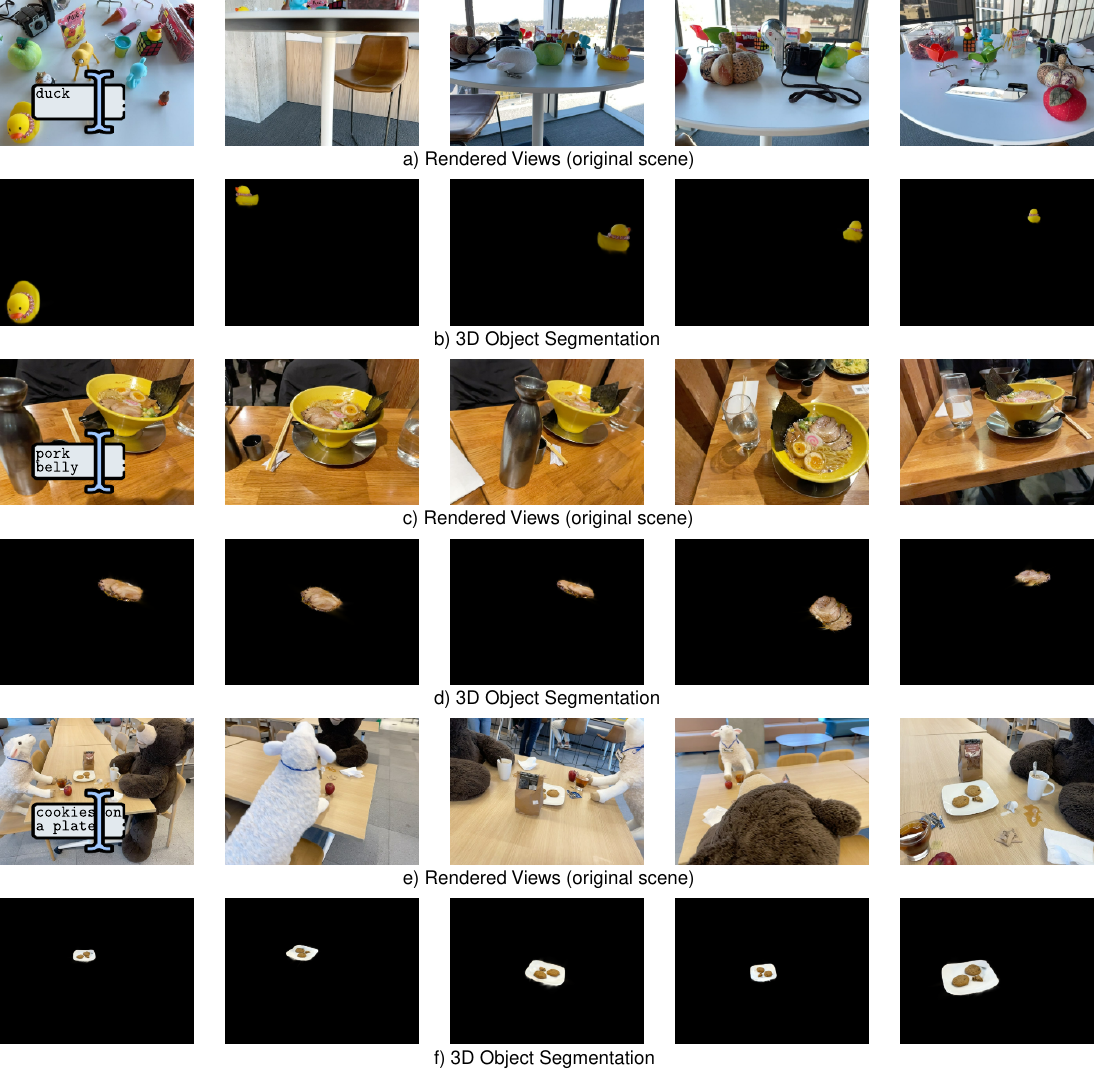}
    \caption{3D Object Segmentation. 
    Rows a), c) and e) present multiple views of the same scene. Given for each scene one text query, we report in b), d), and f) the corresponding view of the 3D object extracted from the 3D Gaussian model of the complete scene. Once we extract all the Gaussians that belong to a 3D object, we are able to visualize it from any point of view. In this figure, you can notice that we are able to see the object even when, in the complete scene, there is an overlap with another object. The most clear example of this is shown in row e), where the \textit{sheep} and the \textit{bear} block the view of the \textit{plate}, but in f) we can see the 3D object since we remove all the Gaussians that do not belong to the \textit{plate}.
    }
    \label{fig:3D_segmentation}
\end{figure}

We here provide further evidence that our model produces consistent segmentations despite inconsistencies across the 2D training segmentation masks. As described in Section $3.2$ of the main paper, we use a text-based query to select a discriminative feature in a reference image, and use it to identify a cluster of 3D Gaussians associated with it and, consequently, with the query. This cluster composes of 3D Gaussians with features having a small cosine distance to the selected discriminative feature, which are then used to compute a convex hull of 3D Gaussians. Finally, we extract all the Gaussians inside this convex hull to achieve 3D object segmentation.
This process is exemplified in Figure~\ref{fig:3D_segmentation}, where we show the 3D object segmentation for various objects in the LERF-Mask dataset, obtained using the same models of Table $1$ of the main paper. By projecting the selected 3D segmentation across views, we observe how the model points to the correct location or the desired object, providing an accurate rendering from the selected viewpoint even in the presence of occlusions, \ie when the object is not visible in the image because of other objects obstructing its line of view.  

\section{Additional Qualitative Results}\label{sec:qualitativelast}
\begin{figure}[t]
    \includegraphics[height=0.8\textheight,trim={0cm 0cm 0cm 0cm}]{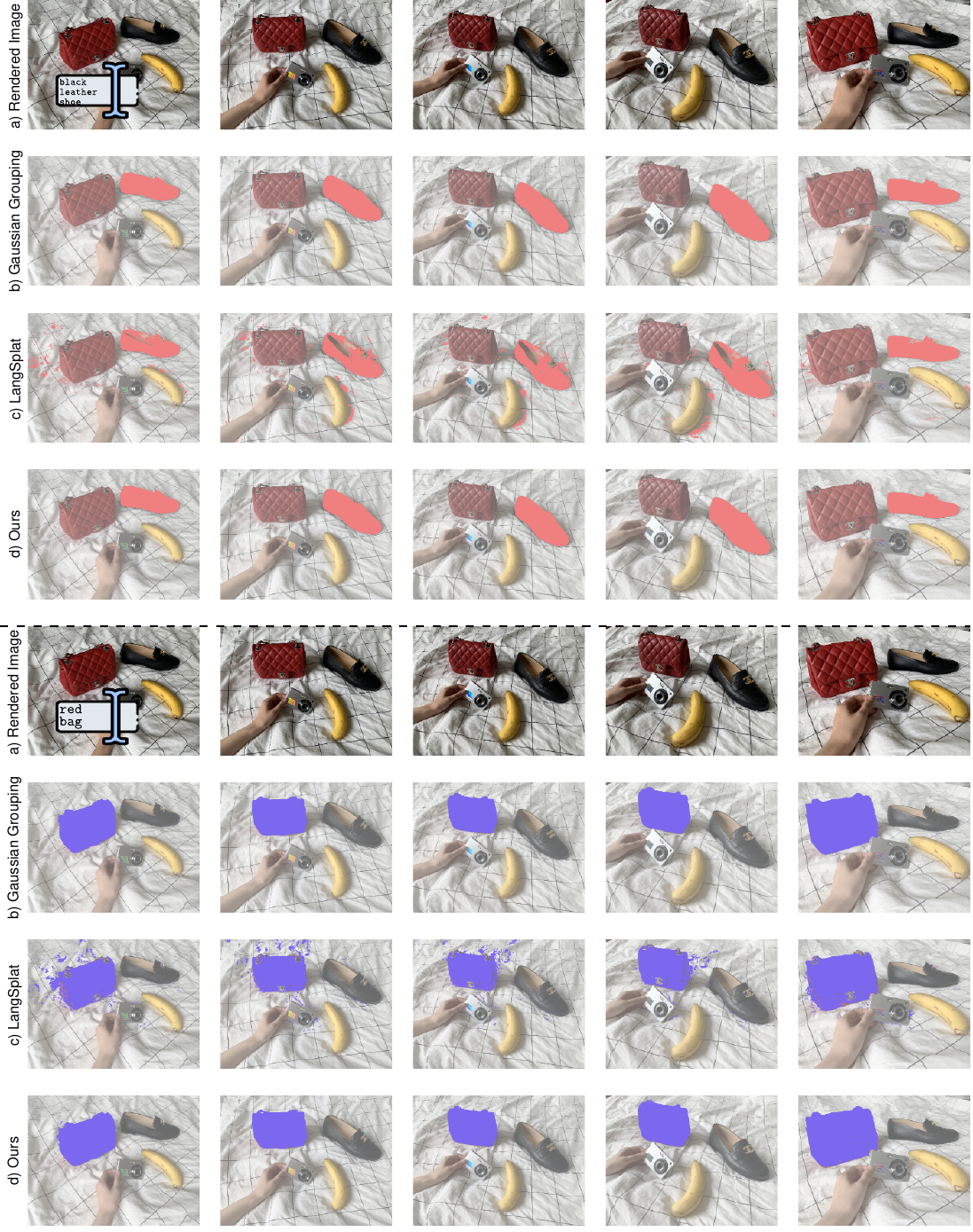}
    \caption{2D mask comparison for LangSplat, Gaussian Grouping, and Contrastive Gaussian Clustering on the \emph{Bed} scene of the 3D-OVS dataset.}
    \label{fig:qualitative_bed}
\end{figure}

\begin{figure}[t]
    \includegraphics[height=0.8\textheight, trim={0cm 0cm 0cm 0cm}]{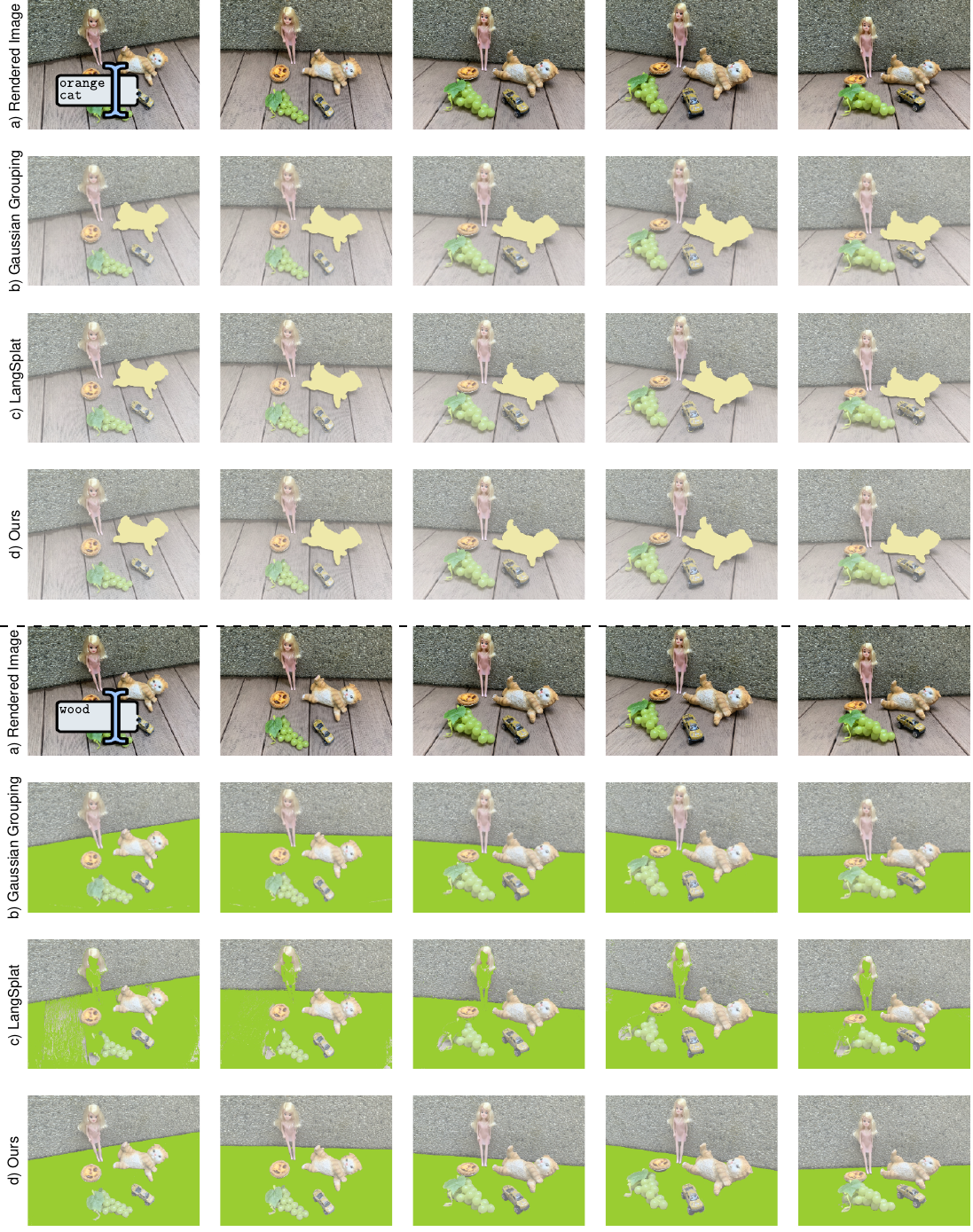}
    \caption{2D mask comparison for LangSplat, Gaussian Grouping, and Contrastive Gaussian Clustering on the \emph{Bench} scene of the 3D-OVS dataset.}
    \label{fig:qualitative_bench}
\end{figure}

\begin{figure}[t]
    \includegraphics[height=0.8\textheight,trim={0cm 0cm 0cm 0cm}]{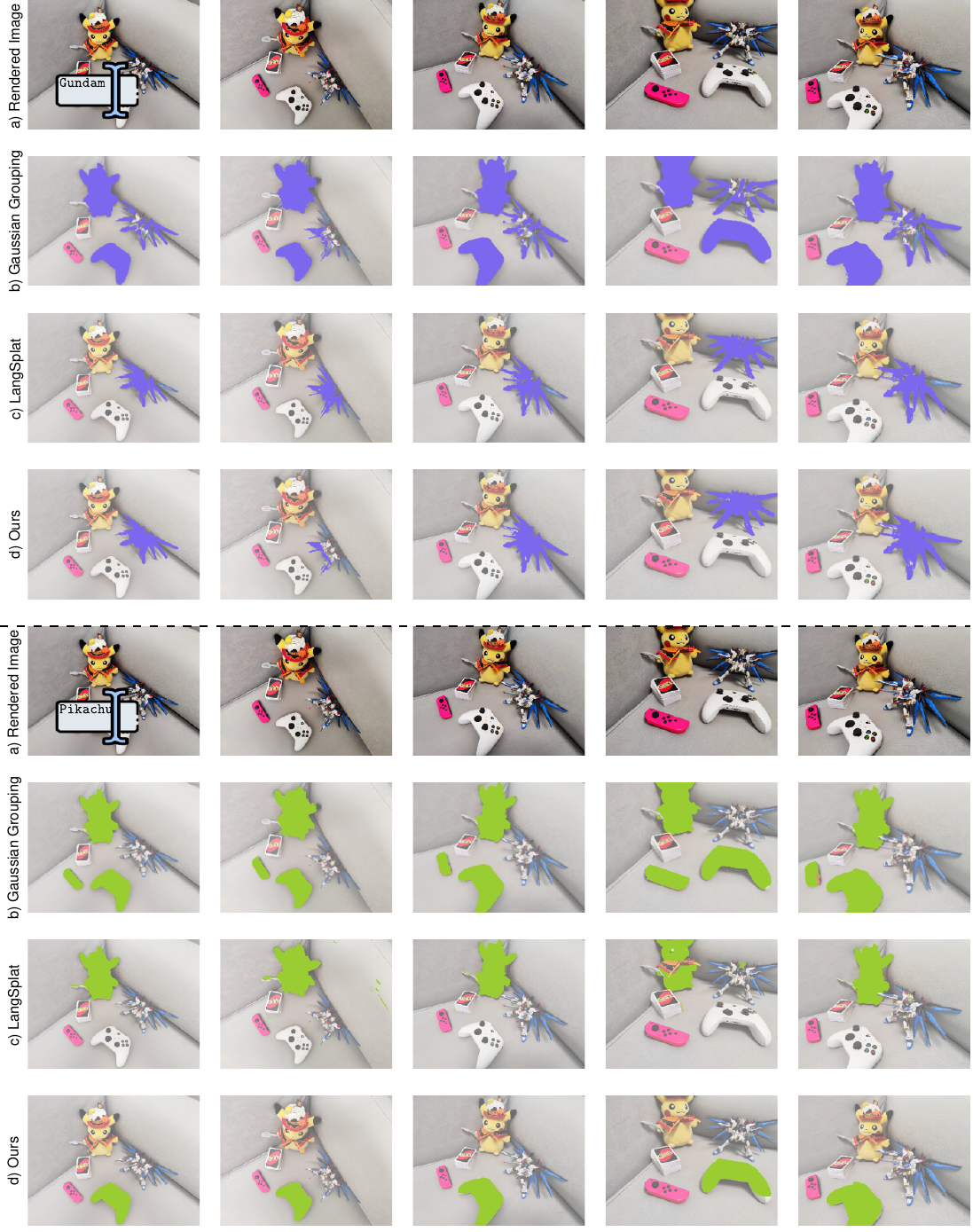}
    \caption{2D mask comparison for LangSplat, Gaussian Grouping, and Contrastive Gaussian Clustering on the \emph{Sofa} scene of the 3D-OVS dataset.}
    \label{fig:qualitative_sofa}
\end{figure}

\begin{figure}[t]
    \includegraphics[height=0.8\textheight,trim={0cm 0cm 0cm 0cm}]{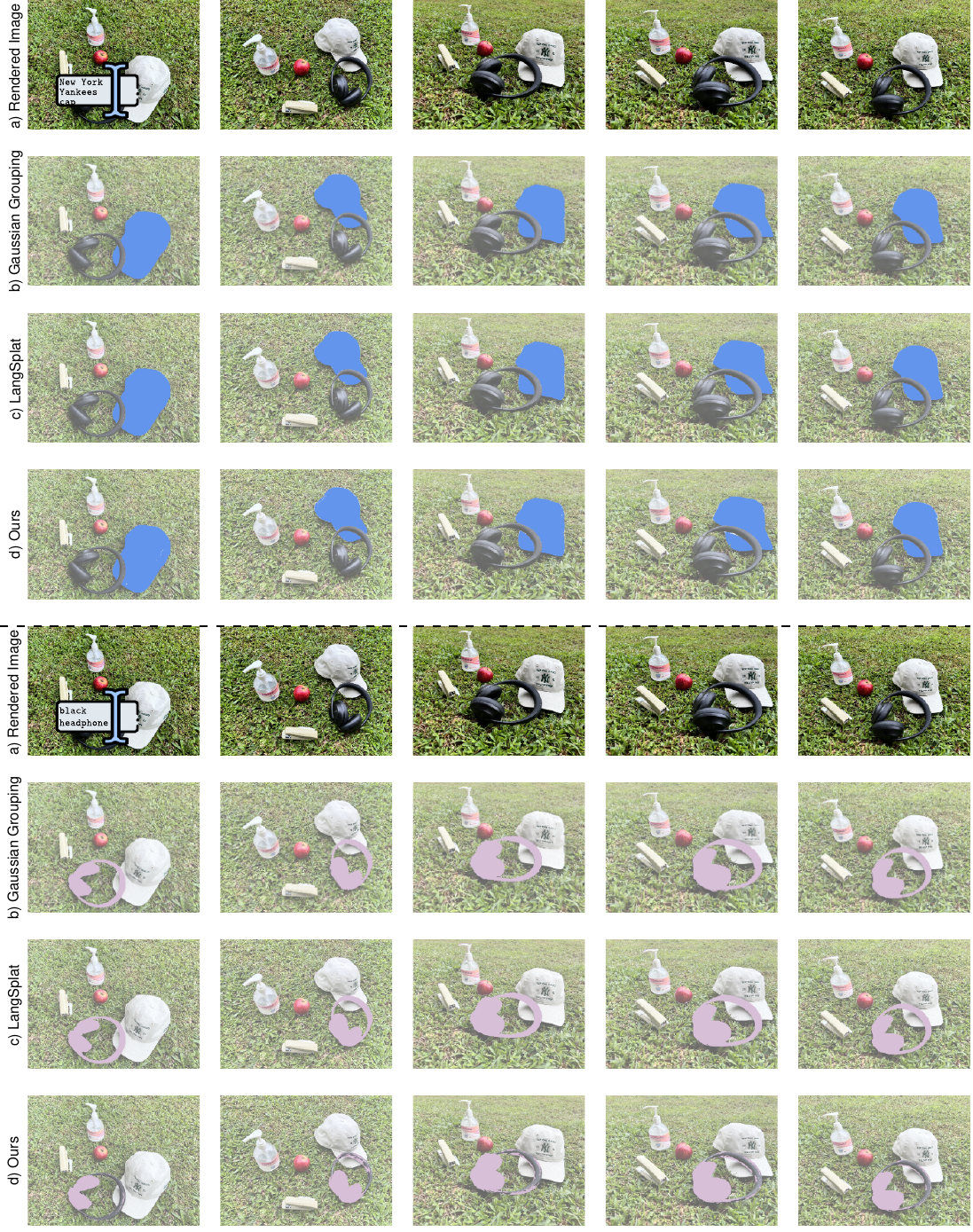}
    \caption{2D mask comparison for LangSplat, Gaussian Grouping, and Contrastive Gaussian Clustering on the \emph{Lawn} scene of the 3D-OVS dataset.}
    \label{fig:qualitative_lawn}
\end{figure}

In this final section, we provide additional insights into the performance of the Contrastive Gaussian Clustering model by comparing its segmentation mask predictions against the masks generated by LangSplat and Gaussian Grouping. We use this also to shed more light on the performance of the model on the 3D-OVS dataset, using the models evaluated in Table 2 of the main paper. Specifically, for each scene of 3D-OVS, we visualize 2D segmentation masks over test-set views for two different queries. Figure~\ref{fig:qualitative_bed}-\ref{fig:qualitative_lawn} showcase such results for the Bed, Bench, Sofa, and Lawn scenes respectively. In Figure~\ref{fig:qualitative_sofa}, we observe how an incorrect localization of \textit{Pikachu} from Grounding DINO leads to an incorrect semantic segmentation. In the same example, we notice that Gaussian Grouping also generates an incorrect semantic segmentation, but its quantitative error is smaller since the object of interest belongs to their predicted segmentation. As we mentioned in the main paper (Sec.$5$, Limitations), our model cannot recover from catastrophic failures in the 2D segmentations from SAM. This is exemplified by Figure~\ref{fig:qualitative_lawn}, where the 2D segmentations from SAM frequently indicate that \textit{black headphones} are composed of two distinct elements, resulting in our model generating inaccurate segmentation of \textit{black headphones}. For text queries like \textit{Pikachu} and \textit{Gundam} (Figure~\ref{fig:qualitative_sofa}), we notice better performance using LangSplat. In contrast, we observe how our method achieves the most accurate segmentation on the \textit{Bed} and \textit{Bench} scenes ( Figure~\ref{fig:qualitative_bed} and~\ref{fig:qualitative_bench}). 

\bibliographystyle{splncs04}
\bibliography{supplementary_material}